\documentclass[twocolumn]{autart}    

\usepackage{numbersUncertaintyAuto}
\AtBeginDocument{\nocite{achemso-control}}

\newcommand{\Tset}{\mathcal{T}}
\newcommand{\Vset}{\mathcal{V}}

\begin{document}

\begin{frontmatter}
		%
\title{Uncertainty quantification in neural network classifiers -- a local linear approach \thanksref{footnoteinfo}}


\thanks[footnoteinfo]{This paper was not presented at any IFAC
	meeting. Corresponding author M.~Malmstr{\"o}m.}

	\author[Linkan]{Magnus Malmstr{\"o}m}\ead{magnus.malmstrom@liu.se},    
	\author[UU]{Isaac Skog}\ead{isaac.skog@angstrom.uu.se},               
	\author[Linkan]{Daniel Axehill}\ead{daniel.axehill@liu.se},   
	\author[Linkan]{Fredrik Gustafsson}\ead{fredrik.gustafsson@liu.se}  
	
	\address[Linkan]{Link{\"o}ping University, Link{\"o}ping, Sweden}  
	\address[UU]{Uppsala University, Uppsala, Sweden}  
	

	\begin{abstract}                          
Classifiers based on neural networks (\nn) often lack a measure of uncertainty in the predicted class. We propose a method to estimate the probability mass function (\pmf) of the different classes, as well as the covariance of the estimated \pmf. First, a local linear approach is used during the training phase to recursively compute the covariance of the parameters in the \nn. Secondly, in the classification phase another local linear approach is used to propagate the covariance of the learned \nn parameters to the uncertainty in the output of the last layer of the \nn. This allows for an efficient Monte Carlo (\mc) approach for: (i) estimating the \pmf; (ii) calculating the covariance of the estimated \pmf; and (iii) proper risk assessment and fusion of multiple classifiers. Two classical image classification tasks, i.e., \mnist, and \cfarTen, are used to demonstrate the efficiency the proposed method.
\end{abstract}
	
\end{frontmatter}


\section{Introduction}
\noindent
In this paper, the problem of quantifying the uncertainty in the predictions from a neural network (\nn) is studied. The uncertainty in the prediction stems from three different sources: errors caused by the optimization algorithm that is used to train the \nn, errors in the data (aleatoric uncertainty), and errors in the model (epistemic uncertainty). In this paper, the focus is on uncertainty from the two latter sources.

In numerous applications,  e.g., image recognition \cite{krizhevsky2012imagenet}, learning properties in atoms \cite{gilmer2017neural}, and various control tasks \cite{li2017deep,karlsson2021speed}, \nn{s} have shown high performance.
Despite their high performance, the use of \nn{s} in safety-critical applications is limited \cite{grigorescu2020survey,bagloee2016autonomous,paleyes2020challenges}. It is partly a consequence of the fact that their predictions usually do not come with any measure of certainty of the prediction, which is crucial to have in a decision-making process in order to know to what degree the prediction can be trusted. Moreover, the quantified measure of uncertainty can be used to detect and remove outliers in the data. Furthermore, it is not possible to fuse the prediction from the \nn with information from other sensors without knowledge about the uncertainty.

Autonomous driving is an example of a safety-critical application in which it is relevant to be able to perform reliable classifications of, e.g., surrounding objects. In particular, this need was highlighted in the fatal Uber accident in 2018 where the lack of reliable classifications of surrounding objects played a role in the development of events that eventually led to the accident \cite{UberFinal}.

The problem to quantify the uncertainty in the prediction of \nn{s}  has lately gained increasing attention, and numerous methods to calculate the uncertainty have been suggested \cite{damour2020underspecification, Ghahramani2015,patel2022accurate,ovadia2019can,lin2022uncertainty}. For a survey of methods see \cite{gawlikowski2021survey}.
The methods suggested in the literature can broadly be divided into one out of two categories.
One category is based on creating an ensemble of predictions from which the uncertainty in the prediction is computed \cite{lakshminarayanan2016simple, Gal2015a, Taye2018,maddox2019simple,osawa2019practical,ayhan2018test,ilg2018uncertainty,carannante2021enhanced}.
In the other category, the \nn structure is extended and the \nn is trained to learn its own uncertainty \cite{charpentier2020posterior, Kendall2017,gustafsson2019dctd, Blundell2015,izmailov2021dangers,eldesokey2020uncertainty}.

Concerning the first category, it has for example been suggested to create an ensemble by training multiple \nn{s}, from whose predictions the uncertainty is computed by \cite{lakshminarayanan2016simple}. Since training a single \nn is often computationally expensive,  this method has high computational complexity. In practice, it is only feasible from a computational perspective to create small ensembles.
To decrease the computational complexity, it was in \cite{Gal2015a, Taye2018} suggested to use already existing regularization techniques (dropout and batch norm) to sample values of the parameters of the \nn from which these ensembles can be created.  Another method to create ensembles is by sampling values of the parameters during the last part of the training phase \cite{maddox2019simple,osawa2019practical}.
So-called test-time data augmentation methods have also been suggested to do perturbation on the test data to create an ensemble of predictions \cite{ayhan2018test}.
Even though the methods in \cite{Gal2015a, Taye2018,maddox2019simple,osawa2019practical,ayhan2018test} do not need multiple models to be trained they require multiple forward passes. Furthermore, they require specially tailored training algorithms and carefully constructed structures of the \nn.

Another limitation of methods relying on creating ensembles is that they have trouble representing the uncertainty caused by the bias in the prediction from a model mismatch. The bias can be caused by an insufficiently flexible model, which could be a result of too high regularization or too low model order.

The problem can be solved by \nn{s} from the second category, i.e., where the structure of the \nn is extended such that it learns its own uncertainty in the prediction.
However, this requires a more intricate \nn structure with tailored loss functions \cite{charpentier2020posterior, Kendall2017, Blundell2015,gustafsson2019dctd}. As a consequence, the training becomes more complex and computationally expensive.
It also makes the methods sensitive to errors caused by the training algorithm, which are not possible to learn. Furthermore, there is also a need for more data to train complex model structures.

In this paper, we address the two limitations of the aforementioned methods using classical local approximations from the area of system identification \cite{lennart1999system}, which is sometimes referred to as the \textit{delta method} \cite{malmstrom2022detection,malmstrom2021, Hannelore2011}.
For regression tasks, the delta method has previously been used to quantify the uncertainty in the prediction of \nn{s}, see e.g., \cite{Hannelore2011, Gene1997,rivals2000construction,malmstrom2021,immer2021improving,deng2022accelerated}, and extended to classification tasks in \cite{malmstrom2022detection}.

\section{Problem formulation and contributions} \label{sec:problem}
Consider the problem of learning a classifier from the training data set
\begin{equation}
  \mathcal{T}\triangleq\{y_n,x_n\}_{n=1}^N
\end{equation}
Here $y_n\in\{1,\ldots,M\}$ is the class labels and $x_n\!\in\!\mathbb{R}^{n_x}$ is the input data of size $n_x$, e.g., pixels in an image. From a statistical point of view, the learning of the  classifier can be seen as a system identification problem where a model $f(x;\theta)$ that predicts the conditional probability mass function (\pmf) $p(y|x)$ of a categorical distribution, are to be identified. That is, the probability for $y=m$ given the input $x$ is modeled as
\begin{equation}
  p(y=m|x;\theta)=f_m(x;\theta),\quad m=1,\ldots,M
\end{equation}
Here $\theta\!\in\!\mathbb{R}^{n_\theta}$ denote the $n_{\theta}$-dimensional parameter vector that parameterize the model. Further, the subscript $m$ denotes the $m$:th element of the vector-valued output of the function.

To ensure that the model $f(x;\theta)$ fulfills the properties associated with \pmf, i.e., $f_m(x;\theta)\geq 0$ $\forall m$ and $\sum_m f_m(x;\theta)=1$, it is typically structured as
\begin{equation}\label{eq:basic model}
  f(x;\theta)=\text{softmax}\left(g(x;\theta)\right)
\end{equation}
where
\begin{equation}\label{eq:softmax}
  \text{softmax}(z)\triangleq \frac{1}{\sum_{m=1}^{M} e^{z_m}}\begin{bmatrix}
                                                                        e^{z_1} \\
                                                                        \vdots \\
                                                                        e^{z_M}
                                                                      \end{bmatrix}
\end{equation}
and $g(x;\theta)$ describes the underlying model of the classifier.

In the case $ g_m(x;\theta)=\theta^\top\phi_m(x)$, where $\phi_m(x)$ denotes, a possible nonlinear, transformation of the input $x$, then the model in (\ref{eq:basic model}) becomes a standard multinomial logistic regression model~\cite{lindholm2022machine}. Furthermore, if the transformation $\phi_m(x)$ is chosen randomly, the model becomes similar to the one used in extreme learning machine classifiers~\cite{huang2006extreme}.

If a \nn is used for classification, then the model  is given by
\begin{subequations}  \label{eq:deepnn}
	\begin{align}
	&h^{(0)} = x, \label{eq:deepnn1} \\
	& a^{(l+1)} =   \begin{pmatrix} h^{(l)} & 1
	\end{pmatrix}^\top W^{(l)} , \quad l = 0, \ldots , L-1,\\
	&h^{(l)}=  \sigma \big(a^{(l)} \big), \quad l = 1, \ldots , L-1, \\
	&g (x; \theta)  = a^{(L)} \label{eq:deepnn2}.
	\end{align}
Here  $\sigma(\cdot)$ denotes the activation function, where the ReLu function $\sigma(z)=\max(0,z)$ is often used. The latent variable $a^{(l)}$ denotes the value of all the nodes in the $l$'th layer of the \nn, and $h^{(l)}$ denotes the transformation using the activation function of the values in all the nodes in the $l$'th layer of the \nn. The parameters of the \nn model consist of all the weights and biases included in the matrices $W^{(L)}, \ldots,W^{(0)}$, i.e.,
\begin{align}
	\theta & = \begin{bmatrix} \text{Vec}(W^{(L)})^\top & \ldots & \text{Vec}(W^{(0)})^\top \end{bmatrix}^\top.
	\end{align}
\end{subequations}
Here $\text{Vec}(\cdot)$ denotes the vectorization operator.
\subsection{Parameter estimation}
For most \nn the number of model parameters $n_\theta > N$ and the model parameters $\theta$ cannot be uniquely identified from the training data $\mathcal{T}$ without some regularization or prior information regarding the parameters. Let $p(\theta)$ denote the prior for the model parameters. The maximum a posteriori estimate of the model parameters is then given by
\begin{equation} \label{eq:map_estimate}
   \hat{\theta}_N =\argmax_{\theta} p(\theta|\mathcal{T})=\argmax_{\theta} L_N(\theta)+ \ln p(\theta),
\end{equation}
where $p(\theta|\mathcal{T})$ denotes the a posteriori distribution of the parameters and
\begin{equation} \label{eq:crossentroAll}
	L_N(\theta)= \sum_{n=1}^N  \ln f_{y_n}(x_n;\theta)
\end{equation}
denotes the cross-entropy likelihood function~\cite{lindholm2022machine}. Here $y_n$ is used as an index operator for the subscript $m$ of $f_m(x;\theta)$.

\subsection{Prediction and classification}
Once the classifier has been learned, i.e., a parameter estimate $\hat{\theta}_N$ has been computed, then for a new input data point $x^\star$ the probability mass function can be predicted as
\begin{equation}
  \hat{p}(y^\star=m|x^\star;\hat{\theta}_N)=f_m(x^\star;\hat{\theta}_N),\quad m=1,\ldots,M
\end{equation}
and the most likely class can be found as
\begin{equation}
  \hat{y}^\star=\argmax_m f_m(x^\star;\hat{\theta}_N).
\end{equation}
Note that, the full \pmf estimate $f(x;\hat{\theta}_N)$ is needed both for temporal fusion using several inputs from the same class and fusion over different classifiers. Furthermore, even small probabilities can pose a large risk, e.g., there might be a pedestrian in front of a car even if another harmless object is more likely according to the classifier. Hence, it is important that the prediction $\hat{p}(y^\star=m|x^\star;\hat{\theta}_N)$ is accurate. However, it is well known that due to, among other things, uncertainties in the parameter estimates $\hat{\theta}_N$ the disagreement between true and estimated \pmf may be significant. Therefore, methods to calibrate the prediction $\hat{p}(y^\star|x^\star;\hat{\theta}_N)$ such that it better matches $p(y^\star|x^\star)$ has been developed.

\subsection{Temperature scaling}\label{sec:temp scaling}
One of the most commonly used methods to calibrate the predicted \pmf is called temperature scaling \cite{guo2017calibration}. In temperature scaling $g(x;\theta)$ is scaled by a scalar quantity $T$ before the normalization by the softmax operator. With a slight abuse of notation, introduce
\begin{equation}\label{eq:temp scaling}
  f(x^\star;\hat{\theta}_N,T)=\text{softmax}\left(g(x^\star;\hat{\theta}_N)/T\right).
\end{equation}
Via the temperature scaling parameter $T$ the variations between the components (classes) in the predicted \pmf can be enhanced or reduced. When $T\rightarrow 0$, then $f(x^\star;\hat{\theta}_N,T) \rightarrow \vec{e}_i$, where $\vec{e}_i$ denotes the $i$:th standard basis vector, thereby indicating that input $x^\star_n$ with total certainty belongs to class $i$. Similarly, when $T\rightarrow \infty$, then $f_m(x^\star;\hat{\theta}_N,T)\rightarrow 1/M$ $\forall m$, thereby indicating that input $x^\star_n$ is equally probable to belong to any of the classes.

Noteworthy is that the temperature scaling is typically done after the parameters $\theta$ have been estimated. For notational brevity, the dependency on the temperature scaling parameter $T$ will only be explicitly stated when temperature scaling is considered.

\subsection{Marginalization of parameter uncertainties}
A more theoretically sound approach to take the uncertainties in the parameter estimate into account is via marginalization of the \pmf with respect to the parameter distribution. That is, an estimate of the \pmf and its covariance are calculated as
\begin{subequations}\
\begin{align}\label{eq:genmarg}
	&f(x^\star|\Tset) \triangleq \int_{\theta} f(x^\star;\theta) p\big(\theta|\Tset\big) d \theta\\
    &P^f\triangleq \int_{\theta} \bigl(f(x^\star;\theta)-f(x^\star|\Tset)\bigr)\bigl(\cdot\bigr)^\top p\big(\theta|\Tset\big) d \theta
	\end{align}
\end{subequations}
From hereon $(x)(\cdot)^\top$ is used as shorthand notation for $xx^\top$. The integral in (\ref{eq:genmarg}) is generally intractable, but can be approximated by Monte Carlo (\mc) sampling as follows
\begin{subequations}\label{eq:genMC}
	\begin{align}
	&\theta^{(k)} \sim p\big(\theta|\Tset\big), \quad k=1,2,\dots,K,\\
	&\hat{f}(x^\star|\Tset) = \frac{1}{K} \sum_{k=1}^K f(x^\star;\theta^{(k)})\\
    &\hat{P}^f  = \frac{1}{K} \sum_{k=1}^K \bigl(f(x^\star;\theta^{(k)})-\hat{f}(x^\star|\Tset)\bigr)\bigl(\cdot\bigr)^\top.
	\end{align}
\end{subequations}
Here $K$ denotes the number of samples used in the \mc sampling.

\subsection{Challenges and contributions}
To realize the \mc scheme in (\ref{eq:genMC}) the posterior parameter distribution $p\big(\theta|\Tset\big)$ must be computed and samples drawn from this high-dimensional distribution. Our contributions are: (i) a local linearization approach that leads to a recursive algorithm of low complexity to compute an approximation of the posterior parameter distribution $p\big(\theta|\Tset\big)$ during the training phase; (ii) a second local linearization approach to reduce the sampling space from $n_{\theta}$ to $M$-dimensional space in the prediction phase; and as a by-product (iii) a low-complexity method for risk assessment and information fusion.

\section{Posterior parameter distribution} \label{sec:parmacov}
Next, a local linearization approach that leads to a recursive algorithm of low complexity to compute an approximation of the posterior parameter distribution $p\big(\theta|\Tset\big)$ during the training phase is presented.

\subsection{Laplace approximation}
Assume the prior distribution for the model parameters to be normal distributed as $p(\theta)=\mathcal{N}(\theta;0,P_0)$, i.e., $l^2$ regularization is used. Then a Laplace approximation of the posterior distribution $p(\theta|\mathcal{T})$ yields that~\cite{Bishop2006}
\begin{equation}\label{eq:laplace approx}
  p(\theta|\mathcal{T})\approx\mathcal{N}(\theta;\hat{\theta}_N,P^\theta_N),
\end{equation}
where
\begin{equation}\label{eq:PN}
  	P^\theta_N=\left(-\frac{\partial^2 L_{N}(\theta)}{\partial \theta^2}\Biggr|_{\theta=\hat{\theta}_N}+P_0^{-1}\right)^{-1}.
\end{equation}
That is, the prior distribution is approximated by a normal distribution with a mean located at the maximum a posteriori estimate and a covariance dependent upon the shape of the likelihood function in the vicinity of the estimate. The accuracy of the approximation will depend upon the amount of information in the training data $\Tset$.

\subsection{Asymptotic distribution}
According to Bernstein-von Mises theorem~\cite{Johnstone2010}, if the true model belongs to the considered model set, the maximum a posteriori estimate $\hat{\theta}$ converge in distribution to
\begin{equation}
\label{eq:thetahat}
\hat{\theta}_N \stackrel{d}{\longrightarrow} \mathcal{N}(\hat{\theta}_N;\theta_\ast,\fisheri^{-1}),
\end{equation}
when the information in the training data $\Tset$ tends to infinity. Here, $\theta_\ast$ denotes the true parameters and
\begin{equation}\label{eq:parmaCovariance}
	\fisheri\triangleq -\text{E}\bigg\{\frac{\partial^2 L_{N}(\theta)}{\partial \theta^2}\bigg\},
\end{equation}
is the Fisher information matrix.
Given the likelihood function in (\ref{eq:crossentroAll}) the Fisher matrix becomes
\begin{subequations} \label{eq:information_with_M}
\begin{equation}
  \fisheri \simeq \sum_{n=1}^{N} \sum_{m=1}^{M}  \eta_{m,n}  \frac{\partial g_m \! (x_n;\theta)}{\partial \theta} \bigg( \! \frac{\partial g_m \!(x_n;\theta)}{\partial \theta} \! \bigg)^{\! \! \top \!}
\end{equation}
where
\begin{equation}\label{eq:eta}
		\eta_{m,n} \triangleq f_m(x_n;\theta)(1-f_m(x_n;\theta)).
\end{equation}
\end{subequations}
See derivations in~\autoref{sec:app_proof}.

\subsection{Recursive computation of covariance}
\noindent
To compute the parameter covariance $P^\theta_N$ defined by \eqref{eq:PN}, the Hessian matrix of the log-likelihood (\Loglike) must be calculated and then inverted. This has a complexity of $\mathcal{O}(N M n_{\theta}^2+n_{\theta}^3)$, which for large $n_\theta$ and $N$ can become intractable. However, by approximating the Hessian matrix of the \Loglike with the Fisher information matrix  as follows
\begin{equation}\label{eq:PN approx}
  	P^\theta_N\approx\left(\mathcal{I}_{\hat{\theta}_N}+P_0^{-1}\right)^{-1},
\end{equation}
the computation can be done recursively and with a complexity of $\mathcal{O}\big(N M n_{\theta}^2+ N M^3 \big)$. To do so, note that the  $\fisheri$ in \eqref{eq:information_with_M} can be written in a quadratic form by defining
\begin{equation} \label{eq:u_derivative}
u_{m,n}  \triangleq   \sqrt{\eta_{m,n}} \frac{\partial g_m(x_n;\theta)}{\partial \theta}\Bigl|_{\theta=\hat{\theta}_N}.
\end{equation}
To compute $u_{mn} \in \mathbb{R}^{n_{\theta}}$ only the gradient of the \Loglike in  \eqref{eq:crossentroAll} is required, which is nevertheless needed for the estimation of $\theta$. Since $\fisheri$, and so also the covariance $P^\theta_N$, can be written in a quadratic form, it is possible to update it recursively as \cite{malmstrom2022detection}
\begin{subequations} \label{eq:recUpdate_all}
	\begin{align} \label{eq:recUpdate}
	K_n &= P^{\theta}_n U_n   \big( I_M  \! +  U^\top_n  P^{\theta}_n U_n\big)^{-1}\\
	P^{\theta}_{n+1}  &= P^{\theta}_n-K_n U^\top_n P^{\theta}_n,
	\end{align}
\end{subequations}
where $I_r$ denotes the identity matrix of size $r$. Here $P^{\theta}_n$ is the parameter covariance for  $n$ measurements, and  $\uun$ is defined as
\begin{align} \label{eq:rec_whatuis}
\uun = \begin{bmatrix} u_{1,n} & \hdots & u_{M,n} \end{bmatrix} \in \mathbb{R}^{ n_{\theta} \times M}.
\end{align}
The recursion is initialized with $P^{\theta}_0=P_0$.

\subsection{Approximating the covariance} \label{sec:high-level}
\noindent
An \nn often has millions of parameters which might result in the amount of data needed to store $P^{\theta}_N$ being larger than the available memory capacity.
A common approach to handle this is to approximate $P^{\theta}_N$ as a block-diagonal matrix \cite{martens2015optimizing}. Another common approach is to use the approximation 
\begin{align}
P^{\theta}_N \approx \begin{bmatrix}
P^{\theta_r}_N & 0 \\ 0 & 0
\end{bmatrix},
\end{align}
where $P^{\theta_r}_N$ denotes the covariance of the estimated parameters $\theta_r$ corresponding to the weights and biases of the $r$ last layers in the \nn \cite{kristiadi2020being,malmstrom2022detection}. 
Depending of the number of included layers, this approximation might be more or less accurate. To compensate for the approximation error when doing the marginalization in (11), a scaling of $P^{\theta_r}_N$ with factor $T_c \geq 1$ can be introduced. The scaling can be estimated from validation data in a similar manner to the temperature scaling $T$ in Sec. \ref{sec:temp scaling}.
\section{Efficient MC sampling} \label{sec:predictCov}
\noindent
With access to the parameter covariance, one can propagate the uncertainty in the parameters to uncertainty in the prediction with the delta method using the principle of marginalization. Plugging in the approximate Gaussian distribution \eqref{eq:thetahat} into \eqref{eq:genmarg} gives
\begin{subequations} \label{eq:normmarg}
\begin{align}
&f(x^\star|\Tset) = \int_{\theta} f(x^\star;\theta) \mathcal{N} \big(\theta; \hat{\theta}_N, P_N^{\theta}\big) d \theta\\
&P^f= \int_{\theta} \bigl(f(x^\star;\theta)-f(x^\star|\Tset)\bigr)\bigl(\cdot\bigr)^\top \mathcal{N} \big(\theta; \hat{\theta}_N, P_N^{\theta}\big) d \theta
\end{align}
\end{subequations}
from which \mc approximation can be performed
\begin{subequations}
	\begin{align}
	&\theta^{(k)}\sim\mathcal{N} \big(\theta; \hat{\theta}_N, P_N^{\theta} \big), \quad k=1,2,\dots,K, \\
	&\hat{f}(x^\star|\Tset) = \frac{1}{K}\sum_{k=1}^K f\big(x^\star;\theta^{(k)}\big)\\
    &\hat{P}^f  = \frac{1}{K} \sum_{k=1}^K \bigl(f(x^\star;\theta^{(k)})-\hat{f}(x^\star|\Tset)\bigr)\bigl(\cdot\bigr)^\top.
	\end{align}
\end{subequations}
This is a feasible solution to the problem, but it comes with a high computational cost since it requires drawing \mc samples from a high-dimensional Gaussian distribution and evaluating the whole network.

\subsection{Marginalization using the delta method} \label{sec:project}
\noindent
The delta method, see e.g., \cite{malmstrom2021,Hannelore2011}, relies on linearization of the nonlinear model $g(x,\theta)$ and provides a remedy to the problem of sampling from the high-dimensional Gaussian distribution. The idea is to project the uncertainty in the parameters to uncertainty in the prediction before the softmax normalization \eqref{eq:softmax}, thereby drastically reducing the dimension of the distribution that must be sampled.
Using the delta method, the uncertainty in the parameters can be propagated to the prediction before the softmax normalization as
\begin{subequations}\label{eq:linearize}
    \begin{equation}
        p(g(x^\star;\theta)|\Tset)\approx\mathcal{N}\big(g(x^\star;\theta);\hat{g}_N, P_N^{g})
    \end{equation}
    where
	\begin{equation}
	  \hat{g}_N =\text{E}\{g(x^\star;\theta)\}\simeq g(x^\star;\hat{\theta}_N)
	\end{equation}
    and
    \begin{equation}
    \begin{split}
	P_N^{g} &= \Cov\{g(x^\star;\theta)\}\\
    &\simeq\bigg(\frac{\partial}{\partial \theta} g(x^\star;\theta)\big|_{\theta=\hat{\theta}_N}\bigg)^\top \! P^{\theta}_N \frac{\partial}{\partial \theta}g(x^\star; \theta)\big|_{\theta=\hat{\theta}_N}.
    \end{split}
\end{equation}
\end{subequations}
Using this Gaussian approximation of the parameter distribution, the \mc approximation of the marginalization integral becomes
\begin{subequations}\label{eq:gmarg}
	\begin{align}
	&g^{(k)}(x^\star) \sim \mathcal{N}\big(g(x^\star,\theta); \hat{g}_N, P_N^{g}), \quad k \!=\!1,2,\dots K\\
	&f^{(k)}(x^\star) = \mathrm{softmax} \big(g^{(k)}(x^\star)\big), \\
	&\hat{f}(x^\star|\Tset)  = \frac{1}{K} \sum_{k=1}^K f^{(k)}(x^\star), \\
    &\hat{P}^f  = \frac{1}{K} \sum_{k=1}^K \bigl(f^{(k)}(x^\star)-\hat{f}(x^\star|\Tset)\bigr)\bigl(\cdot\bigr)^\top.
	\end{align}
\end{subequations}
To summarize, the main idea of the delta method is linearization performed in two steps. First, the parameter uncertainty is computed using \eqref{eq:thetahat}, and second, the uncertainty is propagated to the output of the model by \eqref{eq:linearize}. Hence, the delta method is a local linear approach that gives a linear approximation of a nonlinear model.

\subsection{Fusion}
Suppose there are a set of independent classifiers, each one providing a marginal distribution $\mathcal{N}\big(g_{N,c};\hat{g}_{N,c}, P_{N,c}^{g})$, $c=1,\ldots,C$. Then the predictions (before the softmax normalization) from these classifiers can be fused as follows \cite{Gustafsson2018}
\begin{subequations}
	\begin{align}
	P_{N}^{g} &= \left( \sum_{c=1}^C \big( P_{N,c}^{g}\big)^{-1} \right)^{-1},\\
	\hat{g}_{N} &= P_{N}^{g} \sum_{c=1}^C \big( P_{N,c}^{g}\big)^{-1}\hat{g}_{N,c}.
	\end{align}
\end{subequations}

If a single classifier is used to classify a set of inputs $x^\star_c$, $c=1,\ldots,C$, known to belong to the same class $y^\star$, then these predictions can be fused as follows
\begin{subequations}
	\begin{align}
	P_{N}^{g} &= (H^\top R^{-1} H)^{-1},\\
	 \hat{g}_{N} &=P_{N}^{g} H^\top R^{-1} z
	\end{align}
where
\begin{equation}
  z=\begin{bmatrix}
	  \hat{g}_{N,1}\\ \vdots \\ \hat{g}_{N,C}
	                                \end{bmatrix}\in\mathbb{R}^{C M}\quad H=\begin{bmatrix}
      I_{M} \\
      \vdots \\
      I_{M}
    \end{bmatrix}\in\mathbb{R}^{C M,M}
\end{equation}
and the block $[R]_{i,j}\in\mathbb{R}^{M,M}$, $i,j=1,\ldots,C$, of the covariance matrix is given by
\begin{equation}
  [R]_{i,j}=\frac{\partial}{\partial \theta} g(x^\star_i;\theta)^\top\big|_{\theta=\hat{\theta}_N} \! P^{\theta}_N \frac{\partial}{\partial \theta}g(x^\star_j; \theta)\big|_{\theta=\hat{\theta}_N}.
\end{equation}
\end{subequations}
After fusion, the \mc sampling in \eqref{eq:gmarg} can be applied as before to compute the \pmf estimate.

\subsection{Risk assessment}
Risk assessment can be defined as the probability $r_m$ that $p(y^\star_n=m|x^\star_n)>\gamma_m$. The probability $r_m$ can be estimated from the identified model $f_m(x^\star_n|\Tset)$ as follows
\begin{equation}
\begin{split}
  \hat{r}_m &= \text{Pr}\{f_m(x^\star_n|\Tset)>\gamma_m\}\\
  &\simeq\frac{1}{K} \sum_{k=1}^K \mathbb{1}\big(f^{(k)}_m(x^\star_n)>\gamma_m\big).
\end{split}
\end{equation}
Here $\mathbb{1}(a>b)$ denotes the indicator function which is one if $a>b$ and zero otherwise.

\section{Validation}
Suppose now we have a validation data set $\Vset=\{y^\circ_n,x^\circ_n\}_{n=1}^{N_\circ}$. How can we validate the estimated \pmf $\hat{f}(x^\circ_n|\Tset)$ obtained from \eqref{eq:gmarg}? The inherent difficulty is that the validation data, just as the training data, consists of inputs and class labels, not the actual \pmf. Indeed, there is a lack of unified qualitative evaluation metrics \cite{gawlikowski2021survey}. That being said, some of the most commonly used metrics are classification accuracy, \Loglike, Brier score, and expected calibration error (\ece).

Both the negative \Loglike and the Brier score are proper scoring rules, meaning that they emphasize careful and honest assessment of the uncertainty, and are minimized for the true probability vector \cite{gneiting2007strictly}. 
However, neither of them is a measure of the calibration, i.e., reliability of the estimated \pmf. Out of these metrics, only \ece considers the calibration. Hence, here \ece is the most important metric when evaluating a method used to measure the uncertainty \cite{guo2017calibration,vaicenavicius2019evaluating}.
The calculation of the Brier score and \ece, together with reliability diagrams are described next. They all can be used to tune the temperature scaling $T$ described in Section~\ref{sec:temp scaling}.

\subsection{Brier score}
The Brier score \cite{gneiting2007strictly,wojcik2022slova} corresponds to the least squares fit
\begin{align}
\frac{1}{N_\circ} \sum_{n=1}^{N_\circ}\sum_{m=1}^{M} \bigl(\delta_{m,y_n}-\hat{p}(y^\circ_n=m|x^\circ_n)\bigr)^2,
\end{align}
where $\delta_{i,j}$ denotes the Kronecker delta function.  Furthermore, $\hat{p}(y^\circ_n=m|x^\circ_n)$ denotes a generic \pmf estimate.

\subsection{Accuracy and reliability diagram}
Accuracy and reliability diagrams are calculated as follows. Calculate the $J$ bin histogram defined as
\begin{equation}
	B_j =  \bigg \{ n: \frac{j-1}{J} \leq \max_m \hat{p}(y^\circ_n=m|x^\circ_n) < \frac{j}{J}   \bigg \}
\end{equation}
from the validation data. For a perfect classifier $B_j = \emptyset$ for $j<J$. For a classifier that is just guessing, all sets are of equal size, i.e., $|B_j|=|B_i|$ $\forall i,j$. Note that $\max_m \hat{p}(y^\circ_n=m|x^\circ_n)\geq 1/M$, so the first bins will be empty if $J>M$.

The accuracy of the classifier is calculated by comparing the size of each set with the actual classification performance within the set. That is,
\begin{subequations}
\begin{equation}
\text{acc}(B_j) = \frac{1}{|B_j|} \sum_{n \in B_j} \mathbb{1}\big(\hat{y}^\circ_n = y^\circ_n\big)
\end{equation}
where
\begin{equation}
  \hat{y}^\circ_n=\argmax_m \hat{p}(y^\circ_n=m|x^\circ_n)
\end{equation}
\end{subequations}
A reliability diagram is a plot of the accuracy versus the confidence, i.e., the predicted probability frequency. A classifier is said to be calibrated if the slope of the bins is close to one, i.e., when $\text{acc}(B_j) = (j-0.5)/J$.

\subsection{Confidence and expected calibration error}
Instead of certainty, from hereon the standard, and equivalent, notion of confidence will be used \cite{guo2017calibration,vaicenavicius2019evaluating}.
The mean confidence in a set is denoted $\text{conf}(B_j)$ and is defined as
\begin{align}
\text{conf}(B_j) = \frac{1}{|B_j|} \sum_{n \in B_j} \max_m \hat{p}(y^\circ_n=m|x^\circ_n),
\end{align}
This is a measure of how much the classifier trusts its estimated class labels. In contrast to the accuracy it does not depend on the annotated class labels $y_n$. Comparing accuracy to confidence gives the \ece, defined as
\begin{align}
\text{\ece} =  \sum_{j}^J \frac{1}{|B_j|} |\text{acc}(B_j)-\text{conf}(B_j)|.
\end{align}
A small value indicates that the weight is a good measure of the actual performance.

\begin{figure}[tb!]
	\centering
	\includegraphics[trim={6cm 1cm 25cm 1cm},clip,width=0.21\columnwidth]{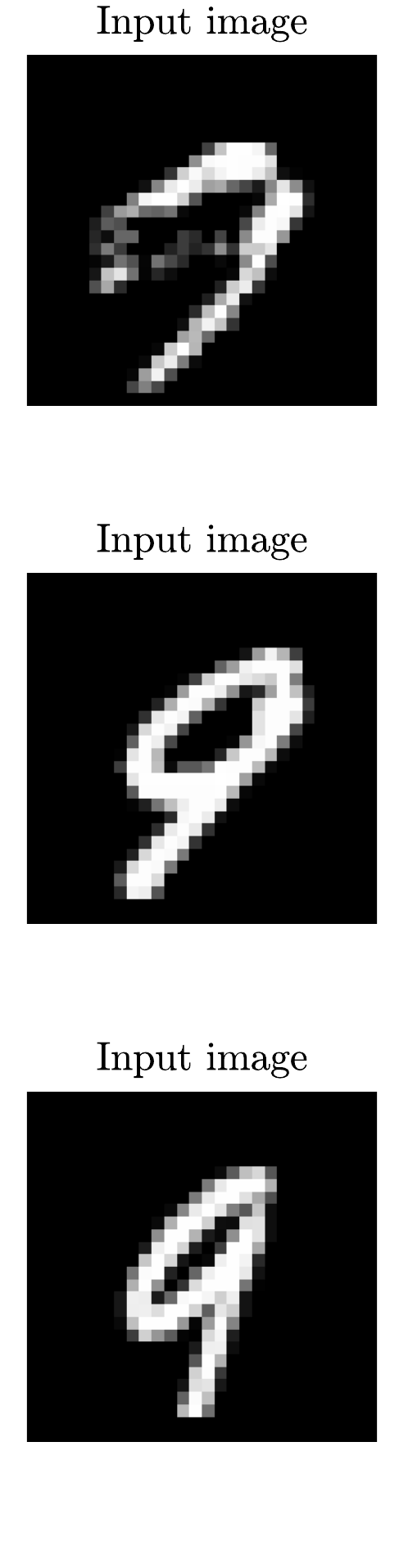}
	\includegraphics[trim={15cm 1cm 3cm 1cm},clip,width=0.76\columnwidth]{fig/covariance_elipse_2023_01_17_b}
	\caption{Example of classification using \eqref{eq:gmarg}. Left: inputs $x^\circ_n$. Middle: Ellipse representation of $P^{g}_N$, \mc samples $g^{(k)}(x^\circ_n)$  and decision line between the classes representing 7 and 9. Right: Estimated \pmf $\hat{f}(x^\circ|\Tset)$.      }
	\label{fig:MM1}
\end{figure}

\section{Experiment study} \label{sec:experiment}
\noindent
To illustrate the application of the proposed method to quantify uncertainty in the prediction, two datasets were used.
First, an \nn was trained using the \mnist dataset \cite{lecun1998mnist} to classify images of handwritten digits. Second, an \nn was trained on the \cfarTen dataset \cite{krizhevsky2009learning} to classify images of ten different objects including e.g., cars, cats, and aircraft.

\subsection{Classification setup}
\noindent
For the two datasets, the structure of the \nn was chosen differently. For the \mnist dataset, a five-layer \nn with fully-connected nodes were used. For the \cfarTen dataset, a LeNet5-inspired structure was used with six convolutional layers followed by four fully connected layers. However, for both datasets the three last layers were chosen to have the same structure, i.e., they were fully connected with $n_{W, L-2} =100$, $n_{W, L-1}=40$, and $n_{W, L}=10$. To decrease the size of the parameter covariance used by the delta method, as described in \autoref{sec:high-level} the first part of the \nn was assumed fixed and used to create high-level features.
Since the structure of the later layers was chosen identically, the parametric models trained on the two datasets had $n_{\bmtheta} = 4450$ parameters.

To estimate the model parameters $\theta$ of the \nn, the \adam optimizer \cite{Kingma2015} was used. The standard \adam optimizer settings, together with an initial learning rate of $10^{-4}$ and $l^2$ regularization of $10^{-4}$, were used. Three and ten epochs were used with the \mnist dataset and \cfarTen dataset, respectively.

\subsection{Illustration of the uncertainties in the predictions}
\noindent
The low-dimensional space of the output from $g(x^\circ_n;\hat{\theta}_N)$ is particularly interesting to study when trying to understand how the uncertainty in the parameter estimate $\hat{\theta}_N$ affects the classification. Even if the parameter covariance $P^{\theta}_N$ is constant and only depends on the training data, the covariance $P^g_N$ depends on the input $x_n$. \autoref{fig:MM1} illustrates this via an example where we concentrate our study on the decision between just a subset of the number of classes in the \mnist dataset, even though the final decision is over all classes. More generally, for some inputs $x^\circ_n$ that are located in a dense region in the space of the training data, the covariance $P^g_N$ is small, but for an input $x^\circ_n$  that is very far from the training data in some norm, the covariance $P^g_N$ can be quite large. This indicates that the parameter estimate is quite sensitive in some directions. That means that the output can also be quite sensitive, and a small change in the parameters can give a completely different output. This can be seen in the two examples on the bottom part of \autoref{fig:MM1}. Even though the estimate of the \pmf looks similar (especially for the two classes under consideration), by studying the unnormalized prediction $g(x^\circ_n;\hat{\theta}_N)$ it is clear that the prediction in the middle has a higher uncertainty compared to the bottom one.

\begin{figure*}[tb!]
\includegraphics[trim={5cm 0cm 4cm 0cm},clip,width=\textwidth]{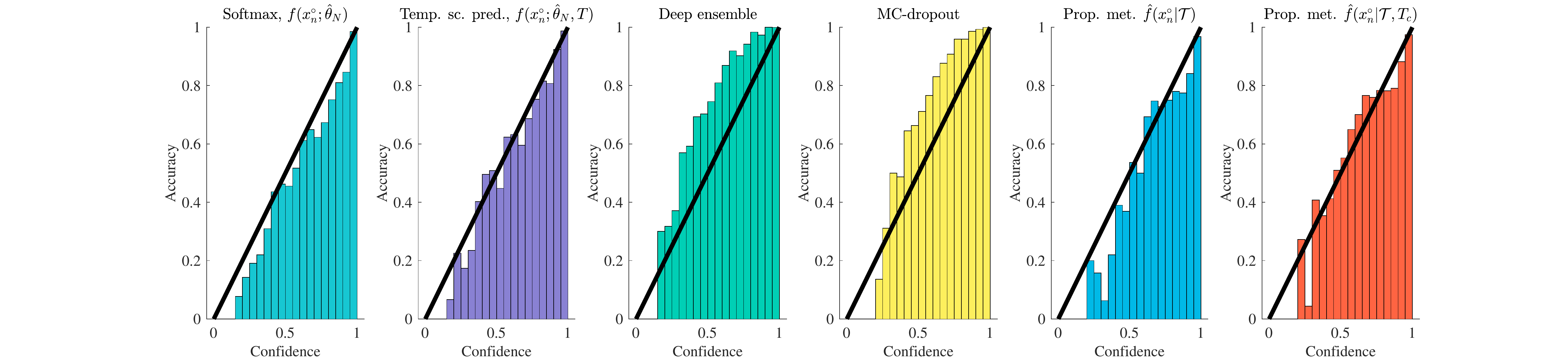}
\caption{ \footnotesize Reliability diagrams for prediction on the \mnist dataset. The diagrams illustrate the six different methods to measure the confidence in the prediction described in \autoref{sec:experiment_result}. A calibration line is also shown in black. }
\label{fig:ReliabilityDiagram_no_extra}
\end{figure*}

\begin{table*}[tb!]
	\centering
	\caption{Computed performance measure for the two datasets. The arrows indicate whether a high or low value is preferable. }
	\begin{tabular}{p{43mm} | c c c c| c c c c }
		\hline
		\hline
		\multicolumn{1}{c|}{} & \multicolumn{4}{c|}{\mnist} & \multicolumn{4}{c}{\cfarTen}\\
		Method & acc. $\uparrow$ & \Loglike $(10^3)$ $\uparrow$ & Brier score $\downarrow$ & \ece $\downarrow$ & acc. $\uparrow$ & \Loglike $(10^3)$ $\uparrow$ & Brier score $\downarrow$ & \ece $\downarrow$ \\ [0.5ex]
		\hline
		Standard $f(x^\circ_n; \hat{\theta}_N)$ & 91$\%$& 7.886 & 0.134 & 1.078 & 83$\%$ &7.904 & 0.291 & 1.328 \\
		Temp. sc. $f(x^\circ_n; \hat{\theta}_N,T)$ & 91$\%$ & 7.818 & 0.133 & 0.951& 83$\%$ & 7.740 & 0.269 &0.612\\
		Deep ensemble & 96$\%$& 7.856 & 0.080 & 2.868 & 87$\%$ &7.834 &0.191 & 1.479\\
		\mc-dropout & 93$\%$& 7.424 & 0.123 & 2.424 & 81$\%$ &9.935 &0.301 & 2.829\\
		Prop. met. $\hat{f}(x^\circ_n| \Tset)$ & 91$\%$ & 7.845 & 0.151 & 1.242 & 83$\%$ &8.176 & 0.243 & 2.140 \\
		Prop. met. $\hat{f}(x^\circ_n| \Tset,T_c)$ & 91$\%$ & 7.763 & 0.151 & 0.821 & 82$\%$ &7.545 & 0.239 &0.540\\ [1ex]
		\hline
		\hline
	\end{tabular}
	\label{table:1}
\end{table*}

\subsection{Results on quantifying the uncertainty} \label{sec:experiment_result}
Six different methods to quantify the uncertainty in the classification, i.e., to estimate $p(y^\circ_n=m|x^\circ_n)$, were evaluated. These are:  
\begin{enumerate}[label=(\roman*)]
\item Standard method, i.e., $\hat{p}(y^\circ_n=m|x^\circ_n)=f_m(x^\circ_n;\hat{\theta}_N)$.
\item Temp. scaling, i.e.,  $\hat{p}(y^\circ_n=m|x^\circ_n)=f_m(x^\circ_n;\hat{\theta}_N,T)$. 
\item Deep ensemble, i.e., $\hat{p}(y^\circ_n=m|x^\circ_n)$ is estimated using the ensemble method in \cite{lakshminarayanan2016simple}; number of trained \nn{s} are 50 for \mnist and 10 for \cfarTen.
\item \mc-dropout, i.e., $\hat{p}(y^\circ_n=m|x^\circ_n)$ is estimated using the ensemble method in \cite{Gal2015a}; $50$ samples of the parameters are used to create the ensemble.
\item Proposed method, i.e., $\hat{p}(y^\circ_n=m|x^\circ_n)=\hat{f}_m(x^\circ_n|\Tset)$.
\item Proposed method with scaled covariance, i.e., $\hat{p}(y^\circ_n=m|x^\circ_n)=\hat{f}_m(x^\circ_n|\Tset,T_c)$, but with the covariance $P_N^{g}$ in \eqref{eq:linearize} scaled with a factor $T_c$.
\end{enumerate}
In \autoref{fig:ReliabilityDiagram_no_extra}, the reliability diagram for the six different methods to quantify the uncertainty in the prediction of the \nn described in \autoref{sec:experiment_result} is shown. Neither computing the uncertainty in the prediction using the softmax (i), deep ensembles (iii), \mc-dropout (iv), or the proposed method without scaled covariance (v) gives calibrated estimates of the uncertainty. To get well-calibrated estimates of the uncertainty either the proposed method with scaled covariance (vi) or temperature scaling (ii) should be used. Finding $T$ and $T_c$ is commonly done by minimizing the \ece. However, increasing the scaling factor decreases the \Loglike. Hence, there is a trade-off between high \Loglike and low \ece. In \autoref{table:1}, the accuracy, \Loglike, Brier score, and \ece are shown for six different methods to quantify the uncertainty in the prediction of the \nn. The methods are evaluated both using the \mnist and \cfarTen datasets. \autoref{table:1} shows that the proposed method attains the lowest \ece for both datasets. This while still having reasonably good performance in terms of accuracy, \Loglike, and Brier score.

\section{Summary and Conclusion} \label{sec:conclusions}
\noindent
A method to estimate the uncertainty in classification performed by a neural network has been proposed.  The method also enables information fusion in applications where multiple independent neural networks are used for classification, or when a single neural network is used to classify a sequence of inputs known to belong to the same class.  The method can also be used for statistical risk assessment.  

The proposed method is based on a local linear approach and consists of two steps. In the first step, an approximation of the posterior distribution of the estimated neural network parameters is calculated. This is done using a Laplacian approximation where the covariance of the parameters is calculated recursively using the structure of the Fisher information matrix. In the second step, an estimate of the PMF is calculated where the effect of the uncertainty in the estimated parameters is considered using marginalization over the posterior distribution of the parameter estimate. This is done by propagating the uncertainty in the estimated parameters to the uncertainty in the output of the last layer in the neural network using a second local linear approach. The uncertainty in the output of the last layer is approximated as a Gaussian distribution of the same dimension as the number of classes. The PMF and its covariance are then calculated via MC sampling, where samples are drawn from this low-dimensional distribution.

The proposed method has been evaluated on two classical classification datasets; MNIST and CFAR10. Neural networks with standard architectures were used. To handle a large number of parameters in these network architectures, only the parameters of the last layer were considered in the uncertainty computations. The results, in terms of ECE, show that the proposed method in its standard form yielded a similar performance as standard methods which do not take the uncertainty in the estimated parameters into account. However, when using a rescaled parameter covariance matrix, used to compensate for the fact that only the uncertainty from the parameters in the last layers was considered, a significant reduction in the ECE was observed. This indicates that the proposed method works, but that more advanced low-rank methods to approximate the parameter covariance are needed.  This is a direction for future research.

\begin{ack}                               
	This work is supported by Sweden's innovation agency, Vinnova, through project iQDeep (project number 2018-02700).
\end{ack}


\bibliography{IEEEabrv,mybibref5G}

\appendix
\section{Derivation of Fisher Information Matrix}  \label{sec:app_proof}  
To calculate the Fisher information matrix in \eqref{eq:information_with_M}, it is necessary to compute the Hessian of the \Loglike with respect to $\theta$. To do so, note that
\begin{equation}
\frac{\partial f_j(x_n,\theta)}{\partial g_i(x_n;\theta)} =f_j(x_n,\theta)(\delta_{i,j}-f_i(x_n,\theta))
\end{equation}
Hence, it holds that 
\begin{align}
\frac{\partial  \ln f_{y_n}(x_n;\theta)}{\partial g(x_n;\theta)} =\vec{e}_{y_n}-f(x_n;\theta).
\end{align}
Using the chain rule the first derivative of the \Loglike (\ref{eq:crossentroAll}) can be computed as
\begin{subequations} \label{eq:firstderivative}
	\begin{align}
	&\frac{\partial L_N(\theta)}{\partial \theta} = \sum_{n=1}^{N}  \frac{\partial g(x_n;\theta)^\top}{\partial \theta} (\vec{e}_{y_n}-f(x_n;\theta))\label{eq:firstderivative_matrix}\\
	& = \sum_{n=1}^{N}\sum_{m=1}^{M} \bigl(\delta_{m,y_n}-f_m(x_n,\theta)\bigr) \frac{\partial g_m(x_n;\theta)}{\partial \theta}.
	\end{align}
\end{subequations}
Differentiation of \eqref{eq:firstderivative} with respect to $\theta$ gives
\begin{subequations}
	\begin{align}
	&\frac{\partial^2 L_{N}(\theta)}{\partial \theta^2} =\sum_{n=1}^{N}\sum_{m=1}^{M} \bigl(\delta_{m,y_n}-f_m(x_n,\theta)\bigr) \frac{\partial^2 g_m(x_n;\theta)}{\partial \theta^2}  \nonumber  \\
	&-  \eta_{m,n} \frac{\partial g_m(x_n;\theta)}{\partial \theta} \bigg( \frac{\partial g_m(x_n;\theta)}{\partial \theta} \bigg)^{\! \top \!}.
	\end{align}
\end{subequations}
with  $\eta_{m,n}$ defined in  \eqref{eq:eta}. And the Fisher information matrix then becomes
\begin{subequations} \label{eq:information_j}
	\begin{align}
	&\fisheri= -\text{E}\bigg\{\frac{\partial^2 \ln L_{N}(\theta)}{\partial \theta^2}\bigg\} \\
	& = -\sum_{n=1}^{N}\sum_{m=1}^{M} \bigl(\text{E}\{\delta_{m,y_n}\}-f_m(x_n,\theta)\bigr)  \frac{\partial^2 g_m(x_n;\theta)}{\partial \theta^2}\nonumber \\
	&+  \eta_{m,n} \frac{\partial g_m(x_n;\theta)}{\partial \theta} \bigg( \frac{\partial g_m(x_n;\theta)}{\partial \theta} \bigg)^{\! \top \!}\\
	&\simeq \eta_{m,n} \frac{\partial g_m(x_n;\theta)}{\partial \theta} \bigg( \frac{\partial g_m(x_n;\theta)}{\partial \theta} \bigg)^{\! \top \!}.
	\end{align}
\end{subequations}
The last approximative equality follows from that $\text{E}\{\delta_{m,y_n}\}=p(y_n=m|x_n)$ and that $f_m(x_n,\theta)$ is an unbiased estimate of $p(y_n=m|x_n)$ when the information in the training data tends to infinity.

\end{document}